\documentclass[11pt,a4paper]{article}
\usepackage[hyperref]{emnlp-ijcnlp-2019}
\usepackage{times}
\usepackage{latexsym}
\usepackage[disable]{todonotes}
\usepackage{xcolor} 
\usepackage{graphicx}
\usepackage{amsmath}
\usepackage{amsfonts}
\usepackage{amssymb}
\usepackage{subfig}
\usepackage{booktabs}
\usepackage{xspace}
\usepackage{arydshln}

\usepackage{url}

\aclfinalcopy

\newcommand\op[1]{\operatorname{#1}}

\newcommand{\fp}{\textsc{FP}\xspace}
\newcommand{\np}{\textsc{NP}\xspace}
\newcommand{\dynsp}{\textsc{DynSp}\xspace}
\newcommand{\camp}{\textsc{Camp}\xspace}
\newcommand{\ours}{\textsc{Ours}\xspace}
\newcommand{\ctx}{\textsuperscript{\textdagger}\xspace}
\newcommand{\tabc}{\textsuperscript{\textasteriskcentered}\xspace}

\title{Answering Conversational Questions on Structured Data\\ without Logical Forms}

\author{Thomas M{\"u}ller, Francesco Piccinno, Massimo Nicosia, Peter Shaw, Yasemin Altun\\
  Google Research \\
  {\tt \{thomasmueller,piccinno,massimon,petershaw,altun\}@google.com} \\}

\date{}

\begin{document}
\maketitle

\begin{abstract}
We present a novel approach to answering sequential questions based on structured objects such as knowledge bases or tables without using a logical form as an intermediate representation. We encode tables as graphs using a graph neural network model based on the Transformer architecture. The answers are then selected from the encoded graph using a pointer network. This model is appropriate for processing conversations around structured data, where the attention mechanism that selects the answers to a question can also be used to resolve conversational references. We demonstrate the validity of this approach with competitive results on the Sequential Question Answering (SQA) task \cite{sqa}.
\end{abstract}

\section{Introduction}
In recent years, there has been significant progress on conversational question answering (QA), where questions can be meaningfully answered only within the context of a conversation \cite{sqa, choi-etal-2018-quac, DBLP:conf/aaai/SahaPKSC18}\todo{Add more citations.}. This line of work, as in single QA setup, falls into two main categories, (i) the answers are extracted from some text in a reading comprehension setting, (ii) the answers are extracted from structured objects, such as knowledge bases or tables. The latter is commonly posed as a semantic parsing task, where the goal is to map questions to some logical form which is then executed over the knowledge base to extract the answers.

In semantic parsing, there is extensive work on using deep neural networks for training models over manually created logical forms in a supervised learning setup \cite{jia2016data, ling2016latent, dong2018coarse}.
However, creating labeled data for this task can be expensive and time-consuming.
This problem resulted in research that investigates semantic parsing with weak supervision where training data consists of questions and answers along with the structured resources
to recover the logical form representations that would yield the correct answer \cite{DBLP:journals/corr/LiangBLFL16,sqa}.

In this paper, we follow this line of research and investigate answering sequential questions with respect to structured objects. In contrast to previous approaches, instead of learning the intermediate logical forms, we propose a novel approach that encodes the structured resources, i.e. tables, along with the questions and answers from the context of the conversation. This approach allows us to handle conversational context without the definition of detailed operations or a vocabulary dependent on the logical form formalism that are required in the weakly supervised semantic parsing approaches. 

We present empirical performance of the proposed approach on the Sequential Question Answering task (SQA) \cite{sqa} which improves the state-of-the-art performance on all questions, particularly on the follow-up questions that require effective encoding of the context.

\section{Approach}
\label{approach}

We build a QA model for a sequence of questions that are asked about a table and can be answered by selecting one or more table cells.

\begin{figure*}
 \centering
 \includegraphics[width=460pt]{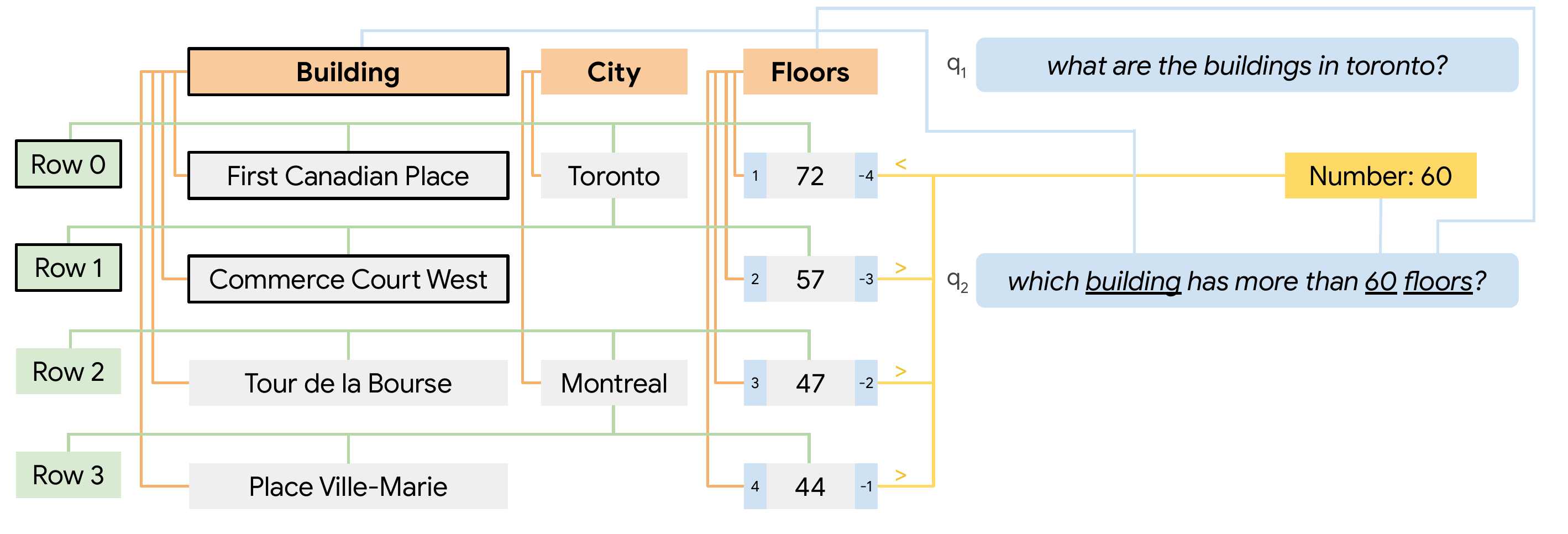}
 \caption{Example of a table encoded as a graph in relation to $q_2$, the follow up question to $q_1$. Columns (orange boxes), rows (green boxes) and cells (gray boxes) are represented by interconnected nodes. Questions words are linked to the table by edit distance (blue lines). Numerical values in question and table are connected by comparison relations (yellow lines). Additionally numerical cells are annotated with their numerical rank and inverse rank with respect to the column (small blue boxes). The answer column, rows and cells of $q_1$ are also marked (nodes with borders).}
 \label{table_example}
\end{figure*}

\subsection{Graph Formulation}
\label{graph_formulation}

We encode tables as graphs by representing columns, rows and cells as nodes. Figure~\ref{table_example} shows an example graph representing how we encode the table in relation to $q_2$, which is the follow up question to $q_1$.
Within a column, cells with identical texts are collapsed into a single node. In the example graph, we only create a single node for \emph{``Toronto''} and a single node for \emph{``Montreal''}.
We then add $4$ directed edges that connect columns and rows to cells, one in either direction (orange and green edges in the figure).

The question is represented by a node covering the entire question text and a node for each token. The main question node is connected to each token, column and cell node.~\footnote{We do not show some of these connections in the figure to avoid clutter.}

Nodes have associated nominal feature sets. All nodes have a feature indicating their type: column, row, cell, question and question token.
The text in column (i.e., the column name), cell, question and token nodes are added to the corresponding node feature set adopting a bag-of-words representation. Column, row and cell nodes have additional features that indicate their column (for cell and column nodes) and row (for cell and row nodes) indexes.

We align question tokens with column names and cell text using the Levenshtein edit distance between n-grams in the question and the table text, similar to previous work \cite{shaw-etal-2019-generating}. In particular, we score every question n-gram with the normalized edit distance~\footnote{$\op{ned}(v,w) = \frac{\op{ed}(v,w)}{\max(|v|,|w|)}$} and connect the cell to the token span if the score is $> 0.5$. Through the alignment, the cell is connected to all the tokens of the matching span and the binned score is added as an additional feature to the cell. In Figure~\ref{table_example}, the \emph{``building''} and \emph{``floors''} tokens in the questions are connected to the matching \emph{``Building''} and \emph{``Floors''} column nodes from the table.

\paragraph{Numeric Operations} In order to allow operations over numbers and date expressions, we extend our graph with a set of relations involving numerical values in the question and table cells. We infer the type of the numerical values in a column, such as the ones in the \emph{``Floors''} column, by picking their most frequent type (number or date). Then, we add special features to each cell in the column: the rank and inverse rank of the value in the cell, considering the other cell values from the same column. These features allow the model to answer questions such as \emph{``what is the building with most floors?''}. In addition, we add  a new node to the graph for each numeric expression from the question (such as the number \textit{60} from the second question in Figure~\ref{table_example}), and we connect it to the tokens it spans. The numerical nodes originated from the question are connected to the table cells containing numerical values. The connection type encodes the result of the comparison between the question and cell values, lesser, greater or equal, as shown in the figure (yellow edges). This relations allow the model to answer questions such as \emph{``which buildings have more than 50 floors?''}. 

\paragraph{Context}
We extend the model to capture conversational context by using the feature-based encoding in the graph formulation. In order to handle follow-up questions, we add the previous answers to the graph by marking all the answer rows, columns and cells with nominal features. 
The nodes with borders in Figure~\ref{table_example} contain the answers to the first question $q_1$: \emph{``what are the buildings in toronto?''}.
In the example, the first two rows receive a feature \textsc{ANSWER\_ROW}, the \emph{``building''} column a feature \textsc{ANSWER\_COLUMN} and \emph{``First Canadian Place''} and \emph{``Commerce Court West''} a feature \textsc{ANSWER\_CELL}.
Notice that the content of $q_1$ is not encoded in the graph, only its answers.

\subsection{Node Representations}

Before the initial encoder layer, all nodes are mapped to vector representations using learned embeddings. For nodes with multiple features, 
such as column and cell nodes, 
we reduce the set of feature embeddings to a single vector using the mean. We also concatenate an embedding encoding whether the node represents a question token, or not.

\subsection{Encoder}

We use a Graph Neural Network (GNN) encoder based on the Transformer~\cite{vaswani2017attention}. The only modification is that the self-attention mechanism is extended to consider the edge label between each pair of nodes. 

We follow the formulation of~\newcite{shaw-etal-2019-generating} that uses additive edge vector representations. The self-attention weights are therefore calculated as:
\begin{equation}
s_{ij} = (\mathbf{W}^{q}x_i)^{\intercal} (\mathbf{W}^{k}x_j + r_{ij}),
\end{equation}
where $s_{ij}$ is the unnormalized attention weight for the node vector representations $x_i$ and $x_j$, and $\mathbf{W}^q$ and $\mathbf{W}^k$ are parameter matrices. This extension introduces $r_{ij}$ to the calculation, which is a vector representation of the edge label between the two nodes.
Edge vectors are similarly added when summing over node representations to produce the new output representations.

We use edge labels corresponding to relative positions between tokens, alignments between tokens and table elements, and relations between table elements, as described in Section~\ref{graph_formulation}. These amount to 9 fixed edge labels in the graph (4 between rows/cells/columns, 2 between question and cells/columns, and 3 numeric relations) and a tunable number of relative token positions).

\subsection{Answer Selection}

\begin{figure}
 \centering
 \includegraphics[width=200pt]{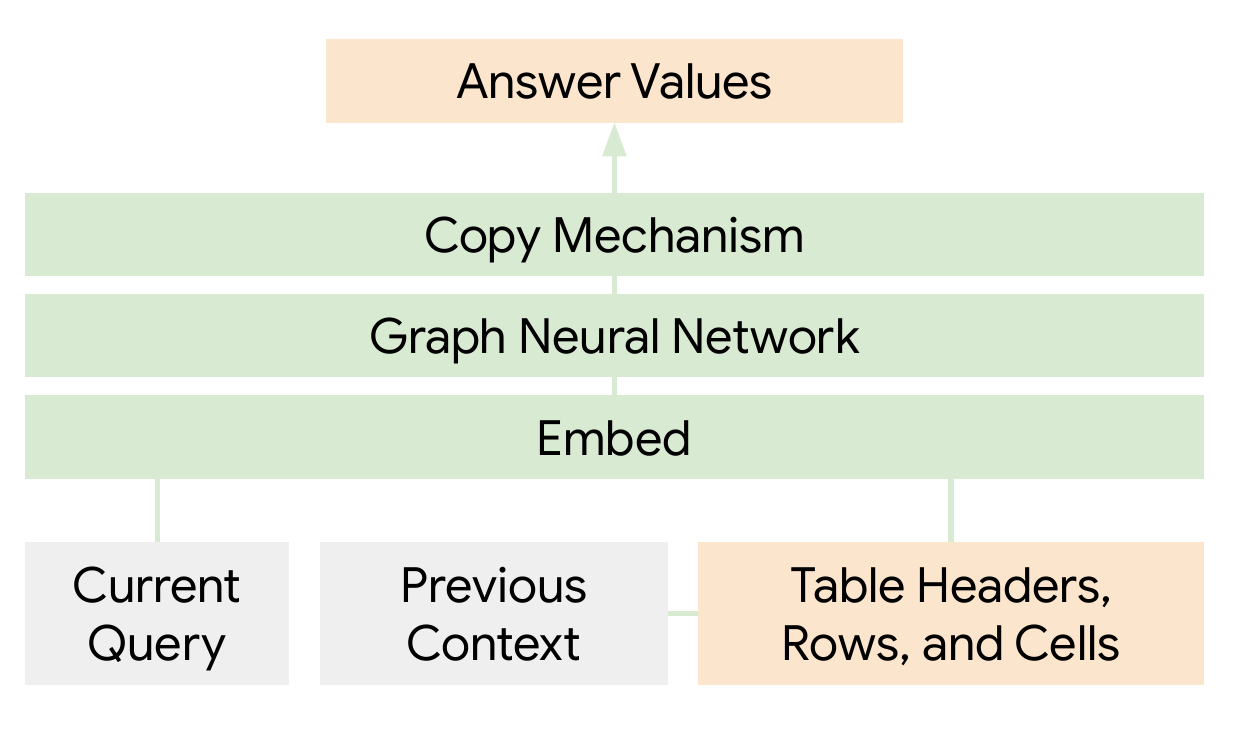}
 \caption{The model input is a graph, with nodes corresponding to query words and table elements, and edge labels capturing their relations. A graph neural network encoder generates contextualized node representations. Answers values are selected from nodes corresponding to table elements.}
 \label{model_diagram}
\end{figure}

We extend the Transformer decoder to include a copy mechanism based on a pointer network~\cite{vinyals2015pointer}. The copy mechanism allows the model to predict sequences of answer columns and rows from the input, rather than select symbols from an output vocabulary.
Figure~\ref{model_diagram} visualizes the entire model architecture.

\section{Related Work}

Semantic parsing models can be trained to produce gold logical forms using an encoder-decoder approach \cite{suhr-etal-2018-learning} or by filling templates \cite{xu2017sqlnet, peng-etal-2017-maximum, yu-etal-2018-typesql}. When gold logical forms are not available, they are typically treated as latent variables or hidden states and the answers or denotations are used to search for correct logical forms \cite{yih-etal-2015-semantic, long-etal-2016-simpler, sqa}. In some cases, feedback from query execution is used as a reward signal for updating the model through reinforcement learning \cite{wikisql, DBLP:journals/corr/LiangBLFL16,mapo2018,merl2019} or for refining parts of the query \cite{execution-guided-neural-program-decoding}. In our work, we do not use logical forms or RL, which can be hard to train, but simplify the training process by directly matching questions to table cells.

Most of the QA and semantic parsing research focuses on single turn questions. We are interested in handling multiple turns and therefore in modeling context. 
In semantic parsing tasks, logical forms \cite{sqa, camp, Guo:2018:DCQ:3327144.3327217} or SQL statements \cite{suhr-etal-2018-learning} from previous questions are refined to handle follow up questions.
In our model, we encode answers to previous questions by marking answer rows, columns and cells in the table, in a non-autoregressive fashion.

In regards to how structured data is represented, methods range from encoding table information, metadata and/or content, \cite{gur-etal-2018-dialsql, camp, petrovski-etal-2018-embedding} to encoding relations between the question and table items \cite{krishnamurthy-etal-2017-neural} or KB entities \cite{sun-etal-2018-open}. We also encode the table structure and the question in an annotation graph, but use a different  modelling approach.

\section{Experimental Setup}
\label{exp_setup}

We evaluate our method on the SequentialQA (SQA) dataset~\cite{sqa}, which consists of sequences of questions that can be answered from a given table.
The dataset is designed so that every question can be answered by one or more table cells. It consists of $6,066$ answer sequences containing $17553$ questions ($2.9$ question per sequence on average). Table \ref{positive_example} shows an example.

We lowercase and remove punctuation from questions, cell texts and column names.
We then split each input utterance on spaces to generate a sequence of tokens.\footnote{Whitespace tokenization simplifies the preprocessing but we can expect an off-the-shelf tokenizer to work as good or even better.} We only keep the most frequent $5,000$ word types and map everything else to one of $2,000$ OOV buckets.
Numbers and dates are parsed in a similar way as in the DynSp and the Neural Programmer \cite{neupro}.

We use the Adam optimizer~\cite{adam} for optimization and tune hyperparameters with  Google Vizier~\cite{vizier}. More details and the hyperparameter ranges can be found in the appendix (\ref{sec:ap_hyper}).

All numbers given for our model are averaged over 5 independent runs with different random initializations.

\section{Results}
\label{sec:results}

\begin{table}
\begin{center}
\scalebox{0.85}{%
\begin{tabular}{lrrrrr}
\toprule
  \textbf{Model} & \textbf{ALL} & \textbf{SEQ} & \textbf{POS1} & \textbf{POS2} & \textbf{POS3} \\
\midrule
  \fp~\tabc            &  34.1  &  7.2 & 52.6 & 25.6 & 25.9 \\
  \np~\tabc            &  39.4  & 10.8 & 58.9 & 35.9 & 24.6 \\
  \dynsp               &  42.0  & 10.2 & {\bf 70.9} & 35.8 & 20.1 \\
\midrule
  \fp~\ctx~\tabc       &  33.2  &  7.7 & 51.4 & 22.2 & 22.3 \\
  \np~\ctx~\tabc       &  40.2  & 11.8 & 60.0 & 35.9 & 25.5 \\
  \dynsp~\ctx          &  44.7  & 12.8 & 70.4 & 41.1 & 23.6 \\
  \camp~\ctx~\tabc     &  45.6  & 13.2 & 70.3 & 42.6 & 24.8 \\ 
\midrule
  \ours~\tabc          &  45.1  & 13.3 & 67.2 & 42.4 & 26.4 \\
  \ours~\ctx~\tabc     &  {\bf 55.1}  & {\bf 28.1} & 67.2 & {\bf 52.7 } & {\bf 46.8}
  \\
\midrule
\midrule
  \ours~\ctx~\tabc (RA)&  61.7  & 28.1 & 67.2 & 60.1 & 57.7 \\
\bottomrule
\end{tabular}
}
\end{center}
\caption{SQA test results. \ctx marks contextual models using the previous question or the answer to the previous question. \tabc marks the models that use the table content. RA denotes an oracle model that has access to the previous reference answer at test time. ALL is the average question accuracy, SEQ the sequence accuracy, and POS X, the accuracy of the X'th question in a sequence.}
\label{test_results}
\end{table}

We compare our model to Float Parser (\fp) ~\cite{floatparser}, Neural Programmer (\np) \cite{DBLP:journals/corr/NeelakantanLAMA16}, \dynsp~\cite{sqa}
and \camp~\cite{camp} 
in Table~\ref{test_results}.

We observe that our model improves the SOTA from $45.6\%$ by \camp to $55.1\%$ in question accuracy (ALL), reducing the relative error rate by $18\%$. For the initial question (POS1), however, it is behind \dynsp by $3.7\%$. 
More interestingly, our model handles follow up questions especially well
outperforming the previously best model \fp by $20\%$ on POS3, a $28\%$ relative error reduction.

As in previous work, we also report performance for a non-contextual setup where follow up questions are answered in isolation. We observe that our model effectively leverages the context information by improving the average question accuracy from $45.1\%$ to $55.1\%$ in comparison to the use of context in \dynsp yielding $2.7\%$ improvement. If we provide the previous reference answers, the average question accuracy jumps to $61.7\%$, showing that $6.6\%$ of the errors are due to error propagation.

\paragraph{Numeric operations}
For understanding how effective our model is in handling numeric operations, we trained models without the specific handling explained in Section~\ref{approach}. We find that that the overall accuracy decreases from $55.1\%$ to $51.5\%$, demonstrating the competence of our approach to model such operations. 
This effectiveness is further emphasized when focusing on questions that contain a superlative (e.g., \emph{``tallest''}, \emph{``most expensive''}) with a performance difference of $47.3\%$ with numeric relations and $40.3\%$ without.
It is worthwhile to call out that the model without special number handling still out-performs the previous SOTA \camp by more than 5 points ($45.6\%$ vs $55.1\%$).

\begin{table*}[!ht]
\begin{center}
\begin{minipage}{.49\linewidth}
\scalebox{0.9}{
\setlength{\arrayrulewidth}{0.75pt}
\begin{tabular}{|l|}
\hline
\bf What are all the nations? \rule{0pt}{3.2ex}\\
\emph{Australia, Italy, Germany, Soviet Union,} \\
\emph{Switzerland, United States, Great Britain, France} \\
\bf Which won gold medals? \rule{0pt}{3.2ex} \\
\emph{Australia, Italy, Germany, Soviet Union} \\
\bf Which won more than one? \rule{0pt}{3.2ex} \\
\emph{Australia} \rule[-2.2ex]{0pt}{0pt} \\
\hline
\end{tabular}
}
\end{minipage}
\begin{minipage}{.49\linewidth}
\scalebox{0.8}{
\begin{tabular}{cccccc}
\toprule
\bf Rank & \bf Nation & \bf Gold & \bf Silver & \bf Bronze & \bf Total \\ 
\midrule
1 & Australia & 2 & 1 & 0 & 3 \\
2 & Italy & 1 & 1 & 1 & 3 \\
3 & Germany & 1 & 0 & 1 & 2 \\
4 & Soviet Union & 1 & 0 & 0 & 1 \\
5 & Switzerland & 0 & 2 & 1 & 3 \\
6 & United States & 0 & 1 & 0 & 1 \\
7 & Great Britain & 0 & 0 & 1 & 1 \\
7 & France & 0 & 0 & 1 & 1 \\
\bottomrule
\end{tabular}
}
\end{minipage}
\end{center}
\caption{A sequence of questions (left) and the corresponding table (right) selected from the SQA dataset that is answered correctly by our approach. This example requires handling conversational context and numerical comparisons.}
\label{positive_example}
\end{table*}

\paragraph{Table size.}
We observe that our model is not sensitive to table size changes, with an average accuracy of $52.4\%$ for the 10\% largest tables (vs. $55.1\%$ overall).\footnote{Fig.~\ref{table_size_plot} of the appendix shows a scatter plot of table size vs. accuracy.}

\paragraph{Error analysis.}

Table \ref{positive_example} shows an example that is consistently handled correctly by the model. It requires a simple string match (\emph{``nations''}$\,\to\,$\emph{``nation''}), and implicit and explicit comparisons.

We performed error analysis on test data over 
$100$ initial (POS$1$) and $100$ follow up questions (POS$> 1$) to identify the limitations of our approach.

For the initial questions, we find that $26\%$ are match errors,
e.g., the model does not match \emph{``episode''} to \emph{``Eps \#''}, or cases where the model has to exclude
rows with empty values from the results.
$29\%$ of the errors require a more sophisticated table understanding,
e.g., rows that represent the total of all other rows should often not be included in the answers.
For $15\%$ of the errors, we think that the reference answer is incorrect and for another $15\%$ the model prediction is correct but contains duplicates because
multiple rows contain the same value. $12\%$ of the errors are around complex matches such as selecting certain ranks (\emph{``the first two''}), exclusion or negation.

For the follow up questions, $38\%$ are caused by complex matches; $17\%$ are match errors; $13\%$ of the errors are due to incorrect reference answers and $11\%$ would require advanced table understanding.
Only $8\%$ of the errors are due to incorrect management of the conversational context.
Section~\ref{sec:app_results} of the appendix contains a more detailed analysis and error examples.

\section{Discussion}

We present a model for table-centered conversational QA that predicts the answers directly from the table. We show that this model improves SOTA on SQA dataset and particularly handles conversational context effectively.

As future work, we plan to expand our model with pre-trained language representations (e.g., BERT \cite{bert}) in order to improve performance on initial queries and matching of queries to table entries. To handle larger tables, we will investigate 
sharding the table row-wise, running the model on all the shards first, and then on the final table which combines all the answer rows.

\bibliography{paper}
\bibliographystyle{acl_natbib}

\appendix

\section{Hyperparameters}
\label{sec:ap_hyper}

We tune hyperparameters with Google Vizier~\cite{vizier}. For the encoder and decoder, we select the number of layers from $[3,6]$ and embedding and hidden dimensions from $\{128, 256, 512\}$, setting the feed forward layer hidden dimensions $4\times$ higher. 
We employ dropout at training time with $P_{dropout}$ selected from $\{0.2, 0.4, 0.5\}$. We select the attention heads from $\{4, 8, 16\}$ and use a clipping distance of $6$ for relative position representations.

We use the Adam optimizer~\cite{adam} with $\beta_1=0.9$, $\beta_2=0.98$, and $\epsilon = 10^{-9}$. We tune the learning rate and use the same warm-up and decay strategy for learning rate as~\citet{vaswani2017attention}, selecting a number of warm-up steps up to a maximum of $2000$. We run the training until convergence for a fixed number of steps ($100,000$) and use the final checkpoint for evaluation. We choose batch sizes from $\{32, 64\}$.

\section{Results}
\label{sec:app_results}

\begin{figure}[!h]
 \centering
 \includegraphics[width=200pt]{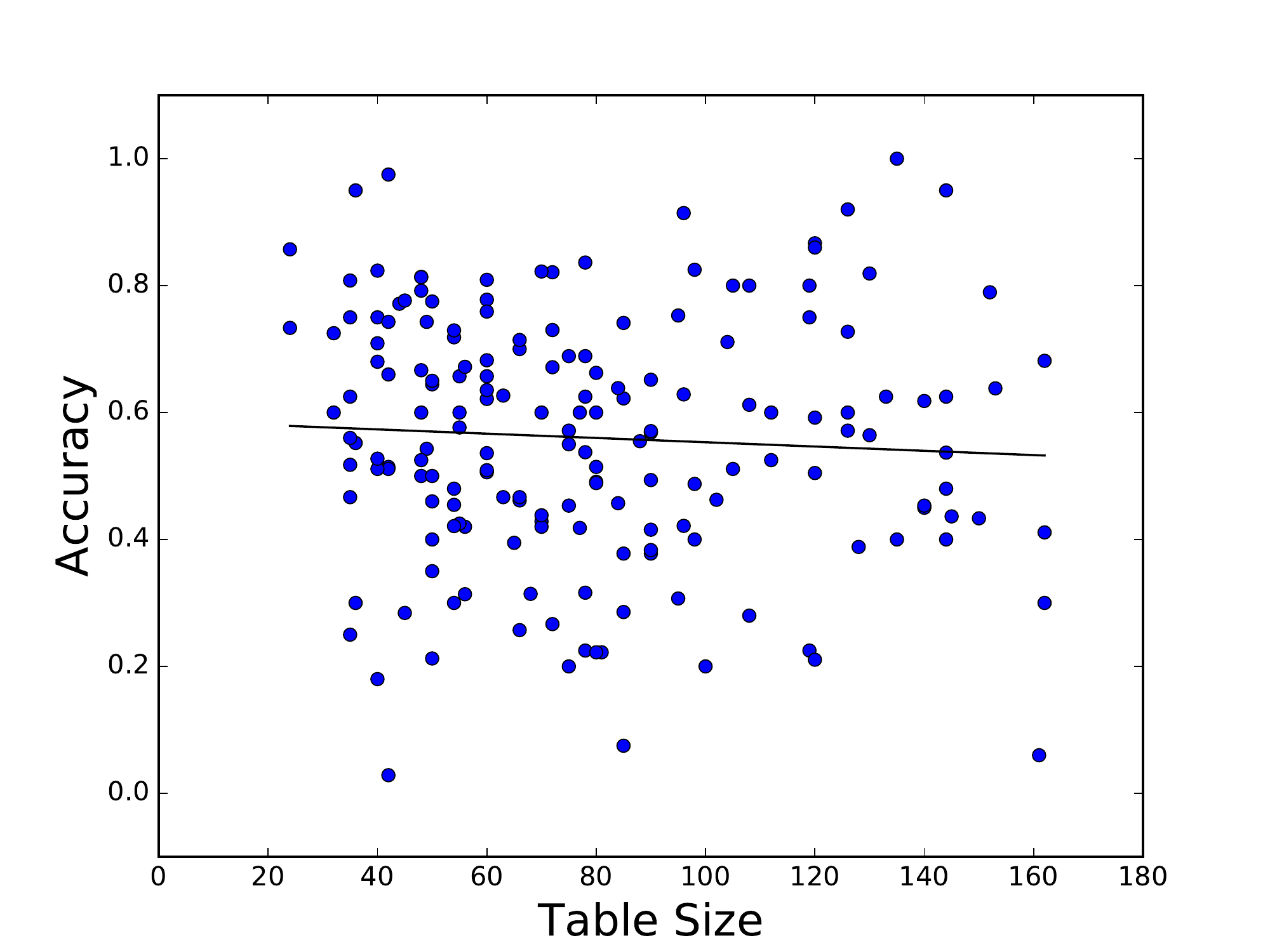}
 \caption{Scatter-plot of accuracy and table size. Each point represents the accuracy on all the questions asked about a test set table of the corresponding size.}
 \label{table_size_plot}
\end{figure}

\paragraph{Table Size}
Given that our model makes use of the whole table, it is conceivable that the performance of our approach can be more sensitive to the table size than methods that predict intermediate representations. Plotting the model performance with respect to number of cells in the table (Figure~\ref{table_size_plot}), we observe that the performance does not vary significantly by the table size.

\begin{table}[!h]
  \centering
  \subfloat[]{
\begin{tabular}{lr}
\toprule
Error type & Counts \\
\midrule
TABLE\_UNDERSTANDING & 29 \\
COMPLEX\_MATCH & 12 \\
MATCH & 26 \\
GOLD & 15 \\
ANSWER\_SET & 15 \\
OTHER & 3 \\
\bottomrule
\end{tabular}
\label{initial}
  }%
  \qquad
  \subfloat[]{
\begin{tabular}{lr}
\toprule
Error type & Counts \\
\midrule
TABLE\_UNDERSTANDING & 11 \\
COMPLEX\_MATCH & 38 \\
MATCH & 17  \\
GOLD & 13 \\
ANSWER\_SET & 4 \\
OTHER & 9 \\
CONTEXT & 8 \\
\bottomrule
\end{tabular}
\label{followup}
  }
  \caption{Errors on 100 random initial \protect\subref{initial} and follow-up \protect\subref{followup} questions}
  \label{tbl:counters}%
\end{table}

\paragraph{Error Analysis}
For a detailed analysis, we annotated $100$ initial and follow-up questions with the following match types:

\paragraph{MATCH} A lexical or semantic match error such as not matching \emph{``episode''} with \emph{``EPS \#''}. 
\paragraph{TABLE\_UNDERSTANDING} A question that would require a higher level table understanding to be answered correctly. For example, we often have to decide that a row is just the total of all other rows and should not be selected.
\paragraph{COMPLEX\_MATCH} A question that would require a numerical value match, some sort of sorting or a negation to be answered. \emph{``what is the area of all of the non-remainders?''}
\paragraph{GOLD} A question with a wrong reference answers. \emph{``what gene functions are listed?''} -- Gold points to Category column rather than Gene Functions.
\paragraph{ANSWER\_SET} The returned answer should be duplicate free. \emph{``what are all of the rider's names?''}, but the table contains \emph{``Carl Fogarty''} multiple times.
\paragraph{CONTEXT} Only used in follow-up questions. This error indicates that a more sophisticated context management is needed.
\paragraph{OTHER} Any other kind of error.

\paragraph{}

Table~\ref{tbl:counters} contains the error counts for initial questions and follow-ups, respectively. Table~\ref{examples} contains interesting examples.

\begin{table*}[!htpb]
\begin{center}
\begin{tabular}[t]{p{200pt}p{200pt}}
\toprule
\bf Question & \bf Notes \\
\midrule
COMPLEX\_MATCH & \\
\midrule
\emph{who scored more than earnie stewart?} & 2-hop reasoning. Requires comparison on top of the result of the inner question.\\
\midrule
\emph{and which has been active for the longest?} & Reasoning with text and date (1986-present) \\
\midrule
\emph{who else is in that field?} & Exclusion. \\
\midrule
\emph{of these, which did not publish on february 9?} & Negation. The model is doing the right thing but missing one of the values.\\
\midrule
\emph{what is the highest passengers for a route?} & The model selects the 2nd highest and not the 1st.\\
\midrule
TABLE\_UNDERSTANDING & \\
\midrule
\emph{now, during which year did they have the worst placement?} & Requires understanding that \emph{``Withdrawal ..''} is worse than any position.\\
\midrule
\emph{which of these seasons have four judges?} & Requires counting the number of named entities within a single cell. \\
\midrule
GOLD & \\
\midrule
\emph{what gene functions are listed?} & Gold points to \emph{``Category''} column rather than \emph{``Gene Functions''}\\
\midrule
\emph{which aircraft have 1 year of service} & Gold points to the 4th column (\emph{``in service''}) instead of the 1st column (\emph{``aircraft''})\\
\midrule
CONTEXT & \\
\midrule
\emph{when was thaddeus bell born?, when was klaus jurgen schneider born?, which is older?} & The correct birthday is selected, but not the person. \\
\bottomrule
\end{tabular}
\end{center}
\caption{Some interesting error cases with comments.}
\label{examples}
\end{table*}

\end{document}